\pdfoutput=1

\documentclass[11pt]{article}

\usepackage[final]{acl}

\usepackage{times}
\usepackage{latexsym}

\usepackage[T1]{fontenc}

\usepackage[utf8]{inputenc}

\usepackage{microtype}

\usepackage{inconsolata}

\usepackage{graphicx}

\usepackage{amsmath,amsfonts,bm}

\def\eqref#1{equation~\ref{#1}}

\def\1{\bm{1}}

\def\vtheta{{\bm{\theta}}}

\DeclareMathAlphabet{\mathsfit}{\encodingdefault}{\sfdefault}{m}{sl}
\SetMathAlphabet{\mathsfit}{bold}{\encodingdefault}{\sfdefault}{bx}{n}

\usepackage{hyperref}
\usepackage{booktabs}
\usepackage{caption}
\usepackage{rotating}
\usepackage{multirow}
\usepackage{enumitem}
\usepackage{amsmath,amsfonts,bm}
\usepackage{extpfeil, extarrows}
\usepackage{xcolor}
\usepackage{multicol}

\newcommand{\tabincell}[2]{\begin{tabular}{@{}#1@{}}#2\end{tabular}}
\newcommand{\M}{\mathcal{M}}
\newcommand{\dtheta}{\Delta \vtheta}

\definecolor{my-yellow}{RGB}{255, 193, 37}
\definecolor{tab-blue}{rgb}{0.12156862745098039, 0.4666666666666667, 0.7058823529411765}
\definecolor{tab-orange}{rgb}{1.0, 0.4980392156862745, 0.054901960784313725}
\definecolor{tab-red}{RGB}{193,0,0}

\title{Model Extrapolation Expedites Alignment}

\author{%
  Chujie Zheng$^{1,2}$\thanks{Work done during Chujie's visit to UCLA. Project repository: \url{github.com/chujiezheng/LLM-Extrapolation}.
  } \quad Ziqi Wang$^{3}$ \quad Heng Ji$^{3}$ \quad Minlie Huang$^{1}$\thanks{Corresponding authors.} \quad Nanyun Peng$^{2}$\footnotemark[2] \\
  $^1$The CoAI Group, DCST, BNRist, Tsinghua University \\
  $^2$University of California, Los Angeles \quad $^3$University of Illinois Urbana-Champaign \\
  \texttt{chujiezhengchn@gmail.com \ aihuang@tsinghua.edu.cn \ violetpeng@cs.ucla.edu} \\
}

\begin{document}
\maketitle
\begin{abstract}
Given the high computational cost of preference alignment training of large language models (LLMs), exploring efficient methods to reduce the training overhead remains an important and compelling research problem.
Motivated by the observation that alignment training typically involves only small parameter changes without injecting new knowledge into models, we propose a straightforward method called \textbf{\textsc{ExPO} ({model extrapolation})} to expedite LLMs' alignment with human preferences.
Given a partially-trained model and its initial SFT checkpoint, \textsc{ExPO} improves the implicit optimization objective of alignment training by simply amplifying the parameter change based on a first-order approximation, without any additional training overhead.
Through controlled experiments, we demonstrate that \textsc{ExPO} boosts a DPO model trained with only 20\% steps to outperform the fully-trained one.
Moreover, we show that \textsc{ExPO} notably improves existing open-source LLMs (ranging from 1.8B to 70B parameters) on the leading AlpacaEval 2.0 and MT-Bench benchmarks, which highlights \textsc{ExPO}'s broader utility in efficiently enhancing LLM alignment.
\end{abstract}

\section{Introduction}
\label{sec:intro}

After conventional unsupervised pre-training on massive textual corpora and supervised fine-tuning (SFT) on high-quality demonstration data, large language models (LLMs) usually require a dedicated training stage to align with human preferences \citep{chatgpt, gpt4, hh-rlhf}, as exemplified by the well-known Reinforcement Learning from Human Feedback (RLHF; \citealt{instructgpt, ppo}) and Direct Preference Optimization (DPO; \citealt{dpo}).
However, alignment training still requires expensive computational resources \citep{ji2024aligner, simpo}, particularly for the larger-sized LLMs (e.g., 70B parameters).
This underscores the significance of exploring more efficient alignment methods to reduce the training overhead.

Our work is first motivated by the observation that preference alignment training typically does not inject new knowledge into models, thereby likely \textit{inducing only small changes of model parameters}.
We support this hypothesis through three arguments.
\textbf{First}, mainstream alignment algorithms like RLHF and DPO incorporate a constraint term (e.g., the KL divergence term) to prevent excessive deviation from the initial SFT checkpoint.
\textbf{Second}, in recent open-source LLM alignment projects \citep{zephyr,tulu,tulu2}, preference alignment training usually adopts smaller learning rates (e.g., 5e-7) and fewer training steps (e.g., 400\textasciitilde500 steps) than SFT.
\textbf{Third}, we take the \texttt{zephyr-7b-dpo} model \citep{zephyr} trained by HuggingFace as a specific instance.
For any two among the pre-trained, SFT, and DPO checkpoints and for any corresponding parameter tensors $\mathbf{P}_1$ and $\mathbf{P}_2$, we compute the Frobenius norm $\left\| \mathbf{P}_1 - \mathbf{P}_2\right\|$ (and a normalized variant)\footnote{
  The Frobenius norm of tensor $\mathbf{P}$ is defined as: $\left\|\mathbf{P}\right\| = \sqrt{ \sum_{i} \mathrm{P}_{i}^2 }$, while the normalized variant is defined as: $\left\|\mathbf{P}\right\| = \sqrt{ \frac{1}{|\mathbf{P}|} \sum_{i} \mathrm{P}_{i}^2 }$, where $|\mathbf{P}|$ denotes the element number of $\mathbf{P}$.
}.
In Table~\ref{tab:norm}, we show that the parameter change of alignment training (i.e., from SFT to DPO) is fairly small, whose absolute value of normalized Frobenius distance is merely $6.348 \times 10^{-6}$, and is also significantly smaller than that of SFT (i.e., from Pre-trained to SFT).
\textbf{Therefore}, in this work we hypothesize that preference alignment training usually involves only small parameter changes.

\begin{table}[htbp]
  \centering
  \caption{Parameter changes of \texttt{zephyr-7b-dpo}.}
  \label{tab:norm}
  \scalebox{0.9}{
      \begin{tabular}{@{\hspace{1mm}}cccc@{\hspace{1mm}}}
          \toprule
          \textbf{CKPT 1} & \textbf{CKPT 2} & \tabincell{c}{\textbf{Frobenius}\\\textbf{Norm}}  & \tabincell{c}{\textbf{Normalized}\\\textbf{Frob Norm}} \\
          \midrule
          Pre-trained & SFT &  $0.9882$  & $1.955 \times 10^{-4}$  \\
          SFT & DPO &  $0.0357$  & $6.348 \times 10^{-6}$  \\
          Pre-trained & DPO &  $0.9889$  & $1.965 \times 10^{-4}$  \\
          \bottomrule
      \end{tabular}
  }
\end{table}

Based on this hypothesis, we formally apply a first-order approximation to the implicit optimization objective of alignment training.
We empirically justify the soundness of this approximation with open-source LLMs, where we show that an interpolated model between the DPO/RLHF model and the initial SFT checkpoint generally exhibits intermediate alignment performance compared to the original models.
Building upon the first-order approximation, we propose a straightforward method called \textbf{\textsc{ExPO} ({model extrapolation})} to expedite LLMs' alignment with human preferences.
\textsc{ExPO} amplifies the parameter change of alignment training to improve the implicit optimization objective, thus bypassing the additional training overhead to achieve better alignment performance.

We conduct controlled experiments to validate \textsc{ExPO}'s effectiveness.
We show that \textsc{ExPO} notably boosts the DPO models using fewer training steps (e.g., only 20\%) to outperform the fully-trained one, with the improvement of up to 8.4\% length-controlled win rate on AlpacalEval 2.0 \citep{alpacaeval}.
We then conduct ablation studies to identify several key factors influencing \textsc{ExPO}'s efficacy, including training data quality, training hyperparameters, and optimizer.
Furthermore, we extend \textsc{ExPO}'s application to twelve open-source LLMs ranging from 1.8B to 70B parameters, which have undergone varied alignment training such as offline DPO, iterative DPO, or online RLHF.
We show that \textsc{ExPO} consistently improves these LLMs by up to 4.5\% on AlpacaEval 2.0 and 0.37 on MT-Bench \citep{llm-as-a-judge}, suggesting that \textsc{ExPO} can also serve as a practical and efficient means to compensate for potential training inadequacy of existing, already-aligned LLMs.
In summary, our work demonstrates the efficacy of model extrapolation in enabling efficient LLM alignment, which can inspire follow-up studies and broader applications in future work.

\section{Methodology}
\label{sec:method}

\subsection{Formulation}

We denote the language model's parameter space as $\bm{\Theta}$ and suppose that the alignment performance can be quantified by a continuous scalar function $\omega: \bm{\Theta} \to \mathbb{R}$, where the higher $\omega(\vtheta)$ indicates the better alignment with human preferences.
In other words, \textit{$\omega(\vtheta)$ is the implicit optimization objective of alignment training}.
Note that $\omega(\vtheta)$ may not have an analytic form.
In practice, we can employ a reward model as a proxy to compare the relative values of $\omega(\vtheta)$ by calculating the expected reward score on a development set of instructions.
We suppose that the model $\M_1$ (parameterized by $\vtheta_1$) has undergone moderate alignment training, and denote its SFT checkpoint as $\M_0$ (parameterized by $\vtheta_0$), which is used for initializing $\M_1$ and satisfies $\omega(\vtheta_0) < \omega(\vtheta_1)$.

\subsection{First-order Approximation}
\label{subsec:first-order}

Based on the aforementioned observation, we suppose that the parameter change from $\M_0$ to $\M_1$, denoted as $\left\| \vtheta_1 - \vtheta_0 \| = \| \dtheta \right\|$, is small.
We can {formally} perform a Taylor Expansion of $\omega$ at $\vtheta_0$ and retain the first-order term:
\begin{align}
    \label{equ:first-order}
    \omega ( \vtheta_0 + \gamma \dtheta ) \approx \omega ( \vtheta_0 ) + \gamma \nabla \omega (\vtheta_0) \cdot \dtheta,
\end{align}
where we define $\gamma \in [0, 1]$ to ensure that $\left\| \gamma \dtheta \right\|$ remains small.
In particular, setting $\gamma=1$ gives:
\begin{align}
    \label{equ:gamma-1}
    &\omega ( \vtheta_1 ) \approx \omega ( \vtheta_0 ) + \nabla \omega (\vtheta_0) \cdot \dtheta, \\
    \label{equ:gamma-1-variant}
    \Longrightarrow \quad &\nabla \omega (\vtheta_0) \cdot \dtheta \approx \omega ( \vtheta_1 ) - \omega ( \vtheta_0 ) > 0.
\end{align}
Thus, the first-order approximation (Equation~\ref{equ:first-order}) essentially predicts that \textit{$\omega ( \vtheta_0 + \gamma \dtheta )$ will improve as $\gamma \in [0, 1]$ increases}.

\begin{figure*}[t]
  \centering
  \includegraphics[width=\linewidth]{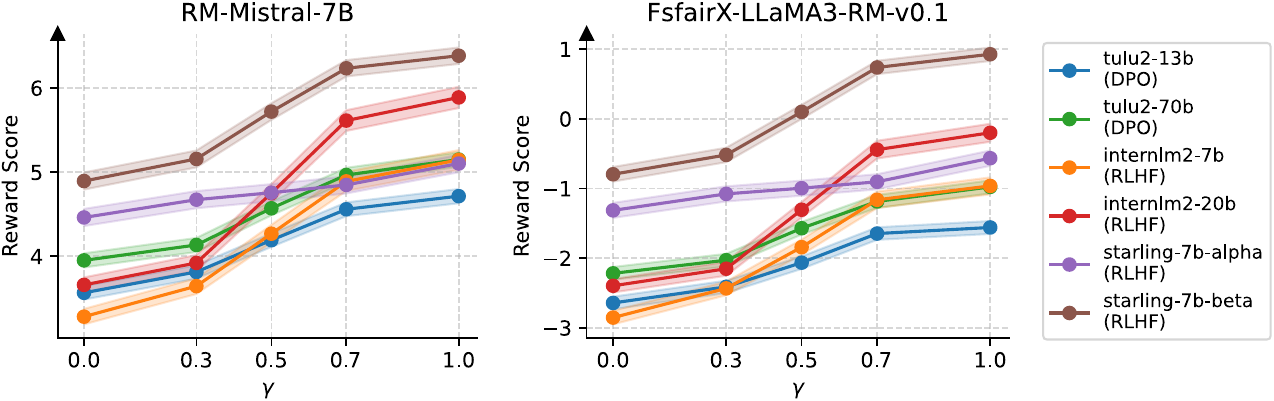}
  \caption{
    Interpolated models usually exhibit intermediate performance between the original DPO/RLHF models and the SFT checkpoints, while their performance improves with increasing $\gamma$ in Equation~\ref{equ:first-order}.
  }
  \label{fig:interpolation}
\end{figure*}

To verify this, we conduct experiments using several open-source DPO/RLHF LLMs \citep{zephyr, internlm2, starling}.
We vary $\gamma$ within $[0, 1]$ and construct interpolated models parameterized by $\vtheta_0 + \gamma \dtheta = (1 - \gamma) \vtheta_0 + \gamma \vtheta_1$.
Their alignment performance is evaluated on the UltraFeedback \citep{ultrafeedback} development set using two open-source reward models: \texttt{RM-Mistral-7B} and \texttt{FsfairX-LLaMA3-RM-v0.1} (detailed experimental setups are described in Section~\ref{subsec:setup}).
Notably, when $\gamma=0$ or $1$, the constructed models degenerate to the original SFT checkpoint $\M_0$ and the DPO/RLHF model $\M_1$, respectively.
The results in Figure~\ref{fig:interpolation} show that the interpolated models constructed via $\vtheta_0 + \gamma \dtheta$ can generate fluent and coherent responses.
Moreover, their alignment performance always lies between the original SFT model $\mathcal{M}_0$ and the DPO/RLHF model $\mathcal{M}_1$, and improves with increasing $\gamma$, which is consistent with the predictions of the first-order approximation.
We thereby empirically justify the soundness of the first-order approximation.

\subsection{\textsc{ExPO}: Model Extrapolation}

In the above first-order approximation, we constrain $\gamma \in [0, 1]$ to maintain the approximation's validity along the straight-line path between $\vtheta_0$ and $\vtheta_1$.
We now consider extending this approximation to the ``extension'' of the line connecting $\vtheta_0$ and $\vtheta_1$ beyond $\vtheta_1$.
Let $\gamma > 1$ and define $\alpha = \gamma - 1 > 0$, denoting $\vtheta_2 = \vtheta_0 + \gamma \dtheta = \vtheta_0 + (1+\alpha) \dtheta$.
By \textit{choosing appropriate $\alpha$ such that $\left\| (1 + \alpha) \dtheta \right\|$ remains small}, we can reformulate the first-order approximation as:
\begin{align}
  \label{equ:first-order-alpha}
  \omega (\vtheta_2)
  \approx &\ \omega ( \vtheta_0 ) + (1+\alpha) \nabla \omega (\vtheta_0) \cdot \dtheta \\
  \notag & \text{(By Equation~\ref{equ:first-order})} \\
  \approx &\ \omega ( \vtheta_1 ) + \alpha \nabla \omega (\vtheta_0) \cdot \dtheta. \\
  \notag & \text{(By Equation~\ref{equ:gamma-1})}
\end{align}
According to Equation~\ref{equ:gamma-1-variant}, we approximately have $\omega(\vtheta_2) > \omega(\vtheta_1)$.
This suggests that, starting from a partially-aligned model $\M_1$ and its SFT checkpoint $\M_0$, by selecting appropriate $\alpha > 0$, we can construct a new model $\M_2$ parameterized by $\vtheta_2$ through amplifying the parameter change $\dtheta$:
\begin{align}
    \label{equ:expo}
    \vtheta_2 = \vtheta_0 + (1+\alpha) \dtheta = \vtheta_1 + \alpha \dtheta,
\end{align}
such that $\M_2$ achieves better alignment performance than $\M_1$.
Consequently, \textit{we improve the implicit optimization objective $\omega(\vtheta)$ of alignment training without requiring additional training}.

\begin{figure}[t]
  \centering
  \includegraphics[width=0.65\linewidth]{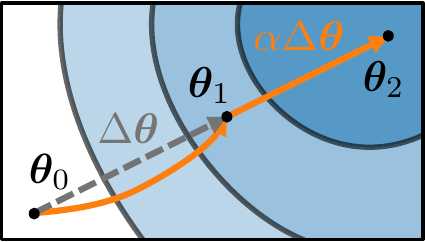}
  \caption{
  The \textbf{\textcolor{tab-orange}{orange} curve} indicates the training trajectory from $\vtheta_0$ to $\vtheta_1$, while the \textbf{\textcolor{tab-orange}{orange} line} denotes the extrapolation from $\vtheta_1$ along \textcolor{gray}{$\dtheta$}, thus producing $\vtheta_2$.
  }
  \label{fig:illustration}
\end{figure}

Since the process of Equation~\ref{equ:expo} essentially ``extrapolates'' the parameters of $\M_1$ along the line connecting $\vtheta_0$ and $\vtheta_1$, we refer to the procedure defined by Equation~\ref{equ:expo} as \textbf{\textsc{ExPO} (model extrapolation)}.
Figure~\ref{fig:illustration} illustrates the \textsc{ExPO} method, where the orange curve from $\vtheta_0$ to $\vtheta_1$ indicates the actual training trajectory from $\M_0$ to $\M_1$, and the straight orange line from $\vtheta_1$ to $\vtheta_2$ denotes the extrapolation from $\M_1$ to $\M_2$.  
In practice, the hyperparameter $\alpha$ in Equation~\ref{equ:expo} (controlling the extrapolation length) can be tuned using inference-level computational resources.
For example, hyperparameter search for a 7B model requires only a single A10 24GB GPU, while a 70B model needs two A100 80GB GPUs.
As high-performance LLM inference frameworks like vLLM \citep{vllm} and SGLang \citep{sglang} continue to rapidly develop, the costs of hyperparameter search will keep decreasing.

\paragraph{Connection to Model Averaging/Interpolation}  
It is worth noting that the idea of ``model averaging'' has been explored in prior work.
Specifically, previous work has discovered that deep neural networks often exhibit mode connectivity \citep{mode-connectivity,linear-mode-connectivity,zhao2020bridging,frankle2020linear}.
This property implies that between two local optima in the parameter space, there typically exists a path where model performance (e.g., validation accuracy or loss) does not degrade significantly during traversal. 
Empirical studies \citep{average-generalization,spurious-feature,model-soups} have shown that even with simple linear \textit{interpolation} paths between two local optima, the loss along the path remains low, and performance often lies between the original models, which is consistent with our observations in Figure~\ref{fig:interpolation}.  
Recent LLM research \citep{mitigating-alignment-tax,super-mario,evolutionary-merging,mergekit} has further explored interpolation across multiple fine-tuned models (i.e., models initialized from the same pre-trained checkpoint but fine-tuned on different data) to create new models with combined capabilities.  
Note that Equation~\ref{equ:expo} can be rewritten as: $\vtheta_2 = (1-\gamma) \vtheta_0 + \gamma \vtheta_1$, which means \textsc{ExPO} can be viewed as a generalized form of model interpolation with weights exceeding 1.  
Hence, the hypothesis we formulated based on the characteristics of preference alignment (i.e., small parameter changes) and the derived \textsc{ExPO} method essentially extend the weight range of traditional model interpolation (from $[0, 1]$ to $(1, +\infty)$).

In the following sections, we will conduct extensive experiments to validate the effectiveness of \textsc{ExPO} in reducing the computational costs of preference alignment training.

\section{Controlled Experiments}
\label{sec:experiments}

\subsection{Setup and Evaluation Protocol}
\label{subsec:setup}

\paragraph{Models and Training Recipe}
Our controlled experiments are based on the \href{https://github.com/huggingface/alignment-handbook/blob/main/recipes/zephyr-7b-beta/dpo/config_full.yaml}{training recipe} of the \texttt{zephyr-7b-dpo} model.
Specifically, we use the UltraFeedback \citep{ultrafeedback} dataset for model training, which contains diverse instruction-response pairs with GPT-4-annotated preference labels and is split into 61K and 1K data as the training and development sets, respectively.
For DPO training, we use \texttt{zephyr-7b-dpo}'s SFT checkpoint for model initialization and as the reference model.
We adopt the global batch size of 128, the learning rate of 5e-7, and the AdamW optimizer \citep{adamw}. %
Note that while \texttt{zephyr-7b-dpo} is trained for 478 steps in total (i.e., one epoch), in \S~\ref{subsec:varying_steps} we will vary the training steps, or equivalently, the training data size.
We train the models on 8 A100 80GB GPUs. %

\paragraph{Inference Details}
We employ the vLLM \citep{vllm} library for high-throughput model inference.
We use top-$k$ ($k=40$) and nucleus sampling \citep{topp} ($p=0.9$) with a temperature of 0.7.
To avoid repetition in generated texts, we set both the factors of presence penalty and frequency penalty to 0.1.
We set the sampling random seed to 42.

\paragraph{Hyperparameter Search}
To determine the optimal $\alpha$ value in \textsc{ExPO}, we use a combination of binary search and grid search with manually tuned intervals (see Appendix~\ref{sec:hyper_search} for details).
We select the $\alpha$ giving the highest expected reward on the UltraFeedback development set (1K instructions), as calculated by the reward model \texttt{RM-Mistral-7B}.

\paragraph{Evaluation Protocol}
We resort to \textbf{AlpacaEval 2.0} \citep{alpacaeval} for model evaluation, which is a leading benchmark that assesses LLMs' instruction-following ability and their alignment with human preferences.
It contains a fixed set of 805 instructions chosen to be representative of real user cases.
For each instruction, it calculates the probability that a GPT-4 Turbo evaluator prefers the output of the evaluated model over the GPT-4 baseline, thus providing an affordable and replicable alternative to human annotation.
The \textbf{win rate} over the GPT-4 baseline is computed as the expected preference probability, while the \textbf{length-controlled (LC) win rate} \citep{alpacaeval-lc} alleviates the length bias of the GPT-4 Turbo evaluator (i.e., the prior preference toward longer responses).

In \S~\ref{subsec:varying_steps}, we report both the raw and LC win rates, as well as the expected reward score over the 805 instructions calculated.
For subsequent experiments, unless otherwise stated, we report the expected reward score on the UltraFeedback development set (1K instructions) for ease of analysis.

\begin{table*}[t]
  \centering
  \caption{Evaluation results on AlpacaEval 2.0 of applying \textsc{ExPO} to DPO models trained with varying steps ($\M_1^{*}$).
  }
  \scalebox{0.9}{
     \begin{tabular}{lccc}
     \toprule
       & \textbf{Reward} & \textbf{Win Rate} & \textbf{LC Win Rate} \\
     \midrule
     SFT ($\M_0$) & 3.42  & 4.7\% & 8.7\% \\
     \cmidrule{1-4}
     DPO, 10\% training steps ($\M_1^{10\%}$)  & 3.97  & 5.9\% & 10.4\% \\
     \quad + \textsc{ExPO} ($\M_2^{10\%}$)  & 6.57 \textcolor{red}{\normalsize (+2.60)}  & 17.9\% \textcolor{red}{\normalsize (+12.0\%)} & 16.3\% \textcolor{red}{\normalsize (+5.8\%)} \\
     \cmidrule{1-4}
     DPO, 20\% training steps ($\M_1^{20\%}$)  & 4.70  & 8.6\% & 12.9\% \\
     \quad + \textsc{ExPO} ($\M_2^{20\%}$)  & \textbf{6.95 \textcolor{red}{\normalsize (+2.25)}}  & \textbf{22.7\% \textcolor{red}{\normalsize (+14.2\%)}} & \textbf{21.3\% \textcolor{red}{\normalsize (+8.4\%)}} \\
     \cmidrule{1-4}
     DPO, 40\% training steps ($\M_1^{40\%}$)  & 5.77  & 12.1\% & 14.6\% \\
     \quad + \textsc{ExPO} ($\M_2^{40\%}$)  & 6.75 \textcolor{red}{\normalsize (+0.98)}  & 17.7\% \textcolor{red}{\normalsize (+5.6\%)} & 16.6\% \textcolor{red}{\normalsize (+2.0\%)} \\
     \cmidrule{1-4}
     DPO, 100\% training steps ($\M_1^{100\%}$)   & 6.16  & 14.7\% & 17.3\% \\
     \quad + \textsc{ExPO} ($\M_2^{100\%}$)  & 6.52 \textcolor{red}{\normalsize (+0.36)}  & 18.0\% \textcolor{red}{\normalsize (+3.3\%)} & 20.2\% \textcolor{red}{\normalsize (+2.8\%)} \\
     \bottomrule
     \end{tabular}%
  }
  \label{tab:varying_steps}%
\end{table*}%

\begin{figure*}[t]
  \centering
  \includegraphics[width=\linewidth]{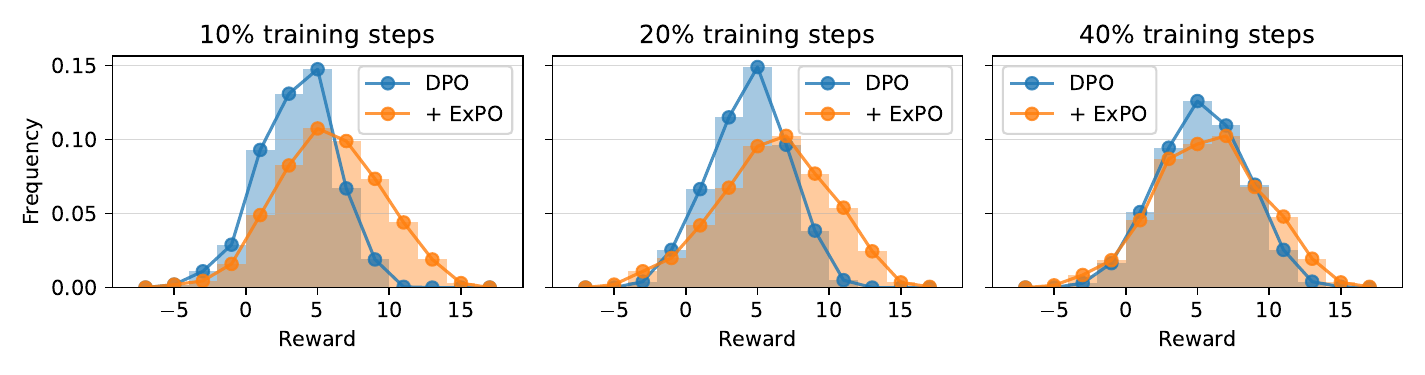}
  \caption{
  Reward distribution on UltraFeedback (development set) for the extrapolated models in Table~\ref{tab:varying_steps}.
  }
  \label{fig:reward_dist}
\end{figure*}

\begin{figure*}[!ht]
  \centering
  \includegraphics[width=\linewidth]{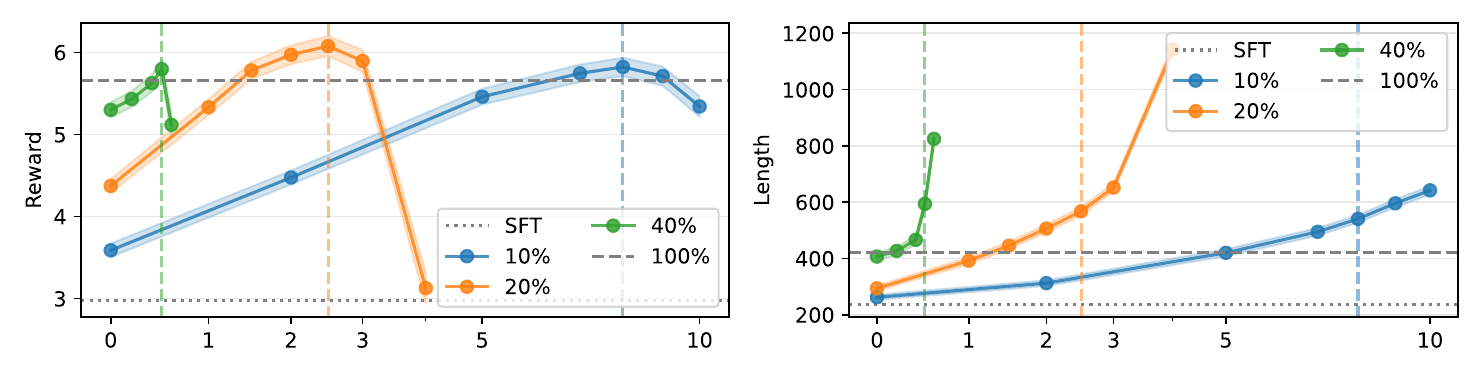}
  \caption{
  $\M_2$'s reward scores and response lengths on UltraFeedback (development set) varying with $\alpha$ (x-axis) for the partially-trained DPO models in \S~\ref{subsec:varying_steps}.
  Dashed vertical lines correspond to the optimal $\alpha$ values.
  $\alpha=0$ indicates that \textsc{ExPO} is not applied (i.e., $\M_1$).
  }
  \label{fig:varying_alpha}
\end{figure*}

\subsection{Analysis of Varying Training Steps}
\label{subsec:varying_steps}

We first investigate \textit{whether \textsc{ExPO} can enhance LLMs with limited alignment training}.
Given that the full training of \texttt{zephyr-7b-dpo} consists of 478 steps (one epoch over the UltraFeedback training data), we initialize from the same SFT checkpoint ($\M_0$) and use the aforementioned training configuration to train DPO models ($\M_1^{*}$) with 10\%, 20\%, and 40\% of the full training steps.
We directly use \texttt{zephyr-7b-dpo} as the 100\%-step (full-training) model $\M_1^{100\%}$.
For these DPO models, we apply \textsc{ExPO} to derive extrapolated models $\M_2^{*}$.

\paragraph{Main Results}
As shown in Table~\ref{tab:varying_steps}, while fewer training steps generally yield lower alignment performance, \textsc{ExPO} effectively bridges the gap caused by reduced training steps.
For example, \textsc{ExPO} boosts $\M_1^{10\%}$'s LC win rate from 10.4\% to $\M_2^{10\%}$'s \textbf{16.3\%} and $\M_1^{20\%}$ from 12.9\% to $\M_2^{20\%}$'s \textbf{21.3\%}, enabling these extrapolated models to match or even surpass the fully-trained $\M_1^{100\%}$.

\paragraph{Hyperparameter Search Analysis}
The optimal $\alpha$ values for $\M_2^{10\%}$, $\M_2^{20\%}$, $\M_2^{40\%}$, and $\M_2^{100\%}$ are 8.0, 2.5, 0.5, and 0.3, respectively.
Figure~\ref{fig:reward_dist} illustrates the reward distributions of these extrapolated models, showing that their response distributions shift toward higher reward regions compared to the original $\M_1^{*}$ models.
In Figure~\ref{fig:varying_alpha}, we show that increasing $\alpha$ within a reasonable range consistently improves alignment performance.
However, excessively large $\alpha$ causes sharp performance drops and abnormal response length increases (e.g., generating gibberish or failing to terminate).
This indicates that overly large $\alpha$ violates the first-order approximation (Equation~\ref{equ:first-order-alpha}) as $\left\|(1+\alpha)\dtheta\right\|$ becomes too large.
Additionally, since more training steps lead to larger $\left\|\dtheta\right\|$, smaller $\alpha$ values are required for models with more training steps (e.g., $\M_1^{100\%}$) to maintain the validity of Equation~\ref{equ:first-order-alpha}, which is consistent with our hyperparameter search results.

\begin{table*}[t]
  \centering
  \caption{Ablation results on UltraFeedback (development set) of adjusting training data quality.
  ``N/A'' denotes that the reward score does not improve after applying \textsc{ExPO} with the smallest $\alpha=0.1$.}
  \scalebox{0.9}{
    \begin{tabular}{lccccc}
    \toprule
    \multirow{2}[2]{*}{\textbf{Training Data}} & \multicolumn{2}{c}{\textbf{Original ($\M_1^{*}$)}} & \multicolumn{3}{c}{\textbf{+ \textsc{ExPO} ($\M_2^{*}$)}} \\
    \cmidrule(lr){2-3} \cmidrule(lr){4-6}
     & \textbf{Reward} & \textbf{Length} & \textbf{Optimal $\alpha$} & \textbf{Reward} & \textbf{Length} \\
     \midrule
    10\% training steps, random ($\M_{*}^{10\%}$) & 3.59  & 262 & 8.0 & \textbf{5.82} & 541 \\
    10\% training steps, length-biased ($\M_{*}^{10\%,\mathrm{b}}$) & 4.62  & 770 & 0.2 & 4.69 & 810 \\
    \cmidrule{1-6}
    20\% training steps, random ($\M_{*}^{20\%}$) & 4.37  & 294 & 2.5 & \textbf{6.08} & 567 \\
    20\% training steps, length-biased ($\M_{*}^{20\%,\mathrm{b}}$) & 5.05  & 748 & 0.4 & 5.11 & 875 \\
    \cmidrule{1-6}
    40\% training steps, random ($\M_{*}^{40\%}$) & 5.30  & 407 & 0.5 & \textbf{5.80} & 594 \\
    40\% training steps, length-biased ($\M_{*}^{40\%,\mathrm{b}}$) & 4.90  & 671 & N/A & N/A & N/A \\
    \bottomrule
    \end{tabular}%
  }
  \label{tab:data_quality}%
\end{table*}%

\paragraph{Computational Cost Analysis}
The fully-trained model $\M_1^{100\%}$ requires about 12 GPU hours (A100 80GB).
In contrast, $\M_2^{20\%}$'s hyperparameter search takes about 0.5 GPU hour, and combined with $\M_1^{20\%}$'s about 2.5-hour training, the total cost is about {3 GPU hours}, leading to a \textbf{75\% reduction} compared to full training while achieving comparable or better alignment performance.
Moreover, \textsc{ExPO}'s hyperparameter search, which only involves model inference, also significantly reduces hardware requirements, e.g., a 7B model requires only a single A10 24GB GPU for search, whereas training typically needs 8 A100 80GB GPUs.
The above results reaffirm the soundness of the first-order approximation and demonstrate \textsc{ExPO}'s effectiveness in reducing computational costs for LLM alignment.

\paragraph{Other Observations}
We also observe two other noteworthy phenomena:
(1) Extrapolated alignment performance does not strictly increase with training steps. For example, $\M_2^{20\%}$ outperforms $\M_2^{100\%}$, suggesting \textsc{ExPO}'s efficacy depends on factors like training data and hyperparameters.
We will explore these factors in \S~\ref{subsec:data_quality} and~\ref{subsec:training_config}.  
(2) Even fully trained models like $\M_1^{100\%}$ benefit from \textsc{ExPO} (LC win rate increases by 2.8\%), indicating that existing already-aligned models may not be fully optimized, and \textsc{ExPO} can fill this gap.
We will apply \textsc{ExPO} to more existing, already-aligned models in \S~\ref{subsec:extension_model}. 

\subsection{Analysis of Training Data Quality}
\label{subsec:data_quality}

\begin{figure}[t]
  \centering
  \includegraphics[width=0.63\linewidth]{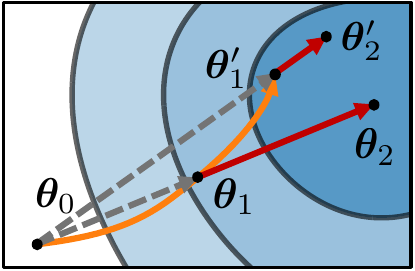}
  \caption{
  Increasing training steps (from $\vtheta_1$ to $\vtheta'_1$) can make the model more prone to learning spurious features from training data, such as length bias.
  This consequently impairs the direction of $\dtheta$ and the achievable performance of \textsc{ExPO} (e.g., $\vtheta'_2$ underperforms $\vtheta_2$).
  }
  \label{fig:illustration_two}
\end{figure}

In the previous section, we observed that alignment performance after model extrapolation does not strictly improve with increased training steps.
We conjecture that this occurs because more training makes the model more prone to learning spurious features from data, such as length bias\footnote{In the UltraFeedback training set, preferred and non-preferred responses have average lengths of 319 and 277 tokens, respectively.} \citep{rdpo}.
According to Equation~\ref{equ:expo}, under our controlled experimental setup where all $\M_1$ are initialized from the same SFT model $\M_0$ and $\vtheta_0$, \textit{the highest achievable performance of the extrapolated model $\M_2$ is uniquely determined by $\dtheta$}.
Hence, \textsc{ExPO}'s effectiveness requires \textbf{$\dtheta$ to indicate the direction that genuinely improves alignment performance}. Learning spurious features like length bias degrades the ``quality'' of $\dtheta$, thus undermining the extrapolation performance.
Figure~\ref{fig:illustration_two} illustrates this phenomenon:
as training steps increase (from $\vtheta_1$ to $\vtheta_1'$), the model can learn spurious features from training data, leading to the degraded alignment performance of extrapolated models (e.g., $\vtheta_2'$ underperforms $\vtheta_2$).

\begin{table*}[t]
  \centering
  \caption{Ablation results of the training epochs, learning rate, and optimizer on UltraFeedback (development set).}
  \scalebox{0.9}{
    \begin{tabular}{llccccc}
      \toprule
      & & \multicolumn{2}{c}{\textbf{Original ($\M_1$)}} & \multicolumn{3}{c}{\textbf{+ \textsc{ExPO} ($\M_2$)}} \\
      \cmidrule(lr){3-4}\cmidrule(lr){5-7}
      &  & \textbf{Reward} & \textbf{Length} & \textbf{Optimal} $\alpha$ & \textbf{Reward} & \textbf{Length} \\
      \midrule
      \multirow{3}[0]{*}{\textbf{Training Epochs}} & 1 (Default) & 4.37  & 294 & 2.5  & \textbf{6.08} & 567 \\
     & 2 ($\times 2$) & 4.93 & 338  & 0.3  & 5.06 & 362  \\
     & 3 ($\times 3$) & 4.47 & 323 & N/A & N/A & N/A \\
     \midrule
     \multirow{3}[0]{*}{\textbf{Learning Rate}} & 5e-7 (Default) & 4.37  & 294 & 2.5  & \textbf{6.08} & 567 \\
      & 1e-6 ($\times 2$) & 5.20 & 374  & 0.5  & 5.54 & 495 \\
      & 2e-6 ($\times 3$) & 5.33 & 365 & 0.4  & 5.52 & 434 \\
      \midrule
      \multirow{3}[0]{*}{\textbf{Optimizer}} & AdamW (Default) & 4.37 & 294 & 2.5  & 6.08 & 567 \\
     & AdaGrad & 3.42 & 246 & 15.0  & 6.25 & 603 \\
     & RMSprop & 4.88 & 344 & 0.4  & 5.08 & 381 \\
     \bottomrule
   \end{tabular}%
  }
  \label{tab:epoch-lr-optim}%
\end{table*}%

To analyze \textit{how training data quality affects \textsc{ExPO}'s effectiveness} in a controlled manner, we take \textbf{length bias} as an example and manually inject length bias into the training data.
Unlike the random sampling in \S~\ref{subsec:varying_steps}, we sort the UltraFeedback training data by the length difference between preferred and non-preferred responses in descending order.
We then train models on the sorted samples orderly so that models will prioritize learning from samples with larger length differences.
From Table~\ref{tab:data_quality}, while introducing length bias temporarily boosts reward scores ($\M_1^{10\%,\mathrm{b}}$ and $\M_1^{20\%,\mathrm{b}}$ outperform $\M_1^{10\%}$ and $\M_1^{20\%}$), extrapolated models consistently underperform ($\M_2^{10\%,\mathrm{b}}$ and $\M_2^{20\%,\mathrm{b}}$ are worse than $\M_2^{10\%}$ and $\M_2^{20\%}$).
Moreover, the optimal $\alpha$ values for $\M_2^{10\%,\mathrm{b}}$ and $\M_2^{20\%,\mathrm{b}}$ are 0.2 and 0.4, which are far smaller than those for $\M_2^{10\%}$ (8.0) and $\M_2^{20\%}$ (2.5).
For $\M_1^{40\%,\mathrm{b}}$, \textsc{ExPO} even fails to yield any improvement.
These results demonstrate that training on biased or low-quality data (e.g., with length bias) causes $\dtheta$ to fail to indicate the direction that genuinely improves alignment performance, thereby diminishing the benefits of model extrapolation.

\subsection{Analysis of Training Configurations}
\label{subsec:training_config}

Next, we analyze \textit{how specific training hyperparameters influence \textsc{ExPO}'s effectiveness}.
Since \textsc{ExPO} amplifies the parameter change $\dtheta$ from $\M_0$ to $\M_1$, we investigate whether \textsc{ExPO} is equivalent to directly increasing the magnitude of parameter changes, such as by raising the \textbf{training epochs} or \textbf{learning rate}.
Additionally, since the training trajectory from $\M_0$ to $\M_1$ (and the resulting $\dtheta$) is closely tied to the gradient descent algorithm, we also explore the impact of the \textbf{optimizer} on \textsc{ExPO}'s effectiveness.
All experiments use the model trained with 20\% steps in \S~\ref{subsec:varying_steps} as the baseline and follow the default training data and configurations.

\paragraph{Training Epochs and Learning Rate}  
We increase the training epochs or learning rate for $\M_1$.
Table~\ref{tab:epoch-lr-optim} shows that while both adjustments improve $\M_1$'s performance, they also reduce the benefits of model extrapolation (lower $\M_2$ performance) and yield smaller optimal $\alpha$ values.
Meanwhile, the $\M_1$ models trained with more epochs or larger learning rates generate significantly longer responses compared to the default setup.
This suggest that both adjustments also make models prone to learning the length bias in training data, thereby degrading $\dtheta$'s quality and the gains from \textsc{ExPO}.  
Notably, when training epochs are set to 3, $\M_1$ cannot benefit from \textsc{ExPO}, likely because the first-order approximation (Equation~\ref{equ:first-order-alpha}) no longer holds as $\left\|\dtheta\right\|$ becomes too large.

\paragraph{Optimizer}  
We train $\M_1$ using three popular optimizers: AdamW~\citep{adamw} (default), AdaGrad~\citep{adagrad}, and RMSprop~\citep{rmsprop}. Table~\ref{tab:epoch-lr-optim} shows that while AdaGrad converges slowest (lowest $\M_1$ performance), it achieves the highest extrapolated alignment performance ($\M_2$), slightly surpassing AdamW.
Conversely, RMSprop, while yielding the best $\M_1$ performance, results in the poorest $\M_2$ performance.
AdamW, as the dominant optimizer in modern LLM training, strikes a balance between convergence efficiency and extrapolated performance.
These results highlight that different optimizers significantly affect $\dtheta$'s quality and extrapolation outcomes.

\begin{table*}[t]
  \centering
  \caption{
     Evaluation results on AlpacaEval 2.0 and MT-Bench of applying \textsc{ExPO} to existing DPO/RLHF LLMs.
  }
  \scalebox{0.9}{
     \begin{tabular}{lcccccc}
     \toprule
        & \multicolumn{3}{c}{\textbf{Original ($\M_1$)}} & \multicolumn{3}{c}{\textbf{+ \textsc{ExPO} ($\M_2$)}} \\
     \cmidrule(lr){2-4} \cmidrule(lr){5-7}
     & \textbf{WR} & \textbf{LC WR} & \textbf{MT-Bench} & \textbf{Win Rate} & \textbf{LC Win Rate} & \textbf{MT-Bench} \\
     \midrule
     \multicolumn{7}{c}{$\M_1$ is trained via \textit{Offline DPO}} \\
     \midrule
      \texttt{zephyr-7b-alpha} & 6.7\% & 10.0\% & 6.85 & \textbf{10.6\% \textcolor{red}{\normalsize (+3.8\%)}} & \textbf{13.6\% \textcolor{red}{\normalsize (+3.6\%)}} & \textbf{6.87 \textcolor{red}{\normalsize (+0.02)}} \\
      \texttt{zephyr-7b-beta} & 10.2\% & 13.2\% & 7.02 & \textbf{11.1\% \textcolor{red}{\normalsize (+0.9\%)}} & \textbf{14.0\% \textcolor{red}{\normalsize (+0.8\%)}} & \textbf{7.06 \textcolor{red}{\normalsize (+0.04)}} \\
      \texttt{tulu2-7b} & 8.5\% & 10.2\% & 6.35 & \textbf{11.5\% \textcolor{red}{\normalsize (+3.0\%)}} & \textbf{11.7\% \textcolor{red}{\normalsize (+1.5\%)}} & \textbf{6.38 \textcolor{red}{\normalsize (+0.03)}} \\
      \texttt{tulu2-13b} & 11.2\% & 15.5\% & 7.00 & \textbf{15.6\% \textcolor{red}{\normalsize (+4.3\%)}} & \textbf{17.6\% \textcolor{red}{\normalsize (+2.1\%)}} & \textbf{7.26 \textcolor{red}{\normalsize (+0.26)}} \\
      \texttt{tulu2-70b} & 15.4\% & 21.2\% & 7.79 & \textbf{23.0\% \textcolor{red}{\normalsize (+7.6\%)}} & \textbf{25.7\% \textcolor{red}{\normalsize (+4.5\%)}} & \textbf{8.03 \textcolor{red}{\normalsize (+0.24)}} \\
     \midrule
     \multicolumn{7}{c}{$\M_1$ is trained via \textit{Iterative DPO}} \\
     \midrule
      \texttt{snorkel-7b-iter} & 24.7\% & 24.0\% & 7.63 & \textbf{28.8\% \textcolor{red}{\normalsize (+4.1\%)}} & \textbf{26.4\% \textcolor{red}{\normalsize (+2.4\%)}} & \textbf{7.69 \textcolor{red}{\normalsize (+0.07)}} \\
      \texttt{llama3-8b-iter} & 29.2\% & 36.0\% & 8.08 & \textbf{32.7\% \textcolor{red}{\normalsize (+3.5\%)}} & \textbf{37.8\% \textcolor{red}{\normalsize (+1.8\%)}} & \textbf{8.45 \textcolor{red}{\normalsize (+0.37)}} \\
     \midrule
     \multicolumn{7}{c}{$\M_1$ is trained via \textit{Online RLHF}} \\
     \midrule
      \texttt{starling-7b-alpha} & 15.0\% & 18.3\% & 7.82 & \textbf{18.2\% \textcolor{red}{\normalsize (+3.2\%)}} & \textbf{19.5\% \textcolor{red}{\normalsize (+1.2\%)}} & \textbf{7.91 \textcolor{red}{\normalsize (+0.09)}} \\
      \texttt{starling-7b-beta} & 26.6\% & 25.8\% & 8.10 & \textbf{29.6\% \textcolor{red}{\normalsize (+3.0\%)}} & \textbf{26.4\% \textcolor{red}{\normalsize (+0.7\%)}} & \textbf{8.18 \textcolor{red}{\normalsize (+0.08)}} \\
      \texttt{internlm2-1.8b} & 3.8\% & 4.0\% & 5.17 & \textbf{5.2\% \textcolor{red}{\normalsize (+1.5\%)}} & \textbf{4.3\% \textcolor{red}{\normalsize (+0.3\%)}} & \textbf{5.26 \textcolor{red}{\normalsize (+0.08)}} \\
      \texttt{internlm2-7b} & 20.5\% & 18.3\% & 7.72 & \textbf{28.1\% \textcolor{red}{\normalsize (+7.6\%)}} & \textbf{22.7\% \textcolor{red}{\normalsize (+4.4\%)}} & \textbf{7.80 \textcolor{red}{\normalsize (+0.08)}} \\
      \texttt{internlm2-20b} & 36.1\% & 24.9\% & 8.13 & \textbf{46.2\% \textcolor{red}{\normalsize (+10.1\%)}} & \textbf{27.2\% \textcolor{red}{\normalsize (+2.4\%)}} & \textbf{8.26 \textcolor{red}{\normalsize (+0.13)}} \\
     \bottomrule
     \end{tabular}%
  }
  \label{tab:extension_model}%
\end{table*}%

\section{Extended Applications of \textsc{ExPO}}
\label{sec:extension}

\subsection{Applying \textsc{ExPO} to More Existing, Already-aligned LLMs}
\label{subsec:extension_model}

In \S~\ref{subsec:varying_steps}, we observed that \textsc{ExPO} also brings noticeable performance improvements to the fully-trained \texttt{zephyr-7b-dpo}.
This motivates us to \textit{apply \textsc{ExPO} to more existing, already-aligned LLMs}.
As hypothesized in \S~\ref{sec:intro}, the normally-trained models should also satisfy the first-order approximation premise, i.e., $\left\|\dtheta\right\|$ is small.
We select twelve open-source models from HuggingFace for experiments (see Appendix~\ref{sec:models} for their model IDs):
\begin{itemize}[leftmargin=*]

\item Five models trained via \textbf{offline DPO}, including \texttt{zephyr-7b-alpha/beta} \citep{zephyr} and \texttt{tulu2-7/13/70b} \citep{tulu};

\item Two models trained via \textbf{iterative DPO}, including \texttt{snorkel-7b-iter} \citep{iterative-dpo} and \texttt{llama3-8b-iter} \citep{rlhflow};

\item Five models trained via \textbf{online RLHF}, including \texttt{starling-7b-alpha/beta} \citep{starling} and \texttt{internlm2-1.8/7/20b} \citep{internlm2}.

\end{itemize}
These models cover a diverse range of model sizes (from 1.8B to 70B) and span three mainstream alignment algorithms widely used in practice.

Based on our hyperparameter search experience for \texttt{zephyr-7b-dpo} in \S~\ref{subsec:varying_steps} (Appendix~\ref{sec:hyper_search}), for the twelve models above, we conduct a simple grid search for the optimal $\alpha$, using the interval of 0.1 within [0.1, 0.5].
In addition to AlpacaEval 2.0, we also evaluate these models on \textbf{MT-Bench} \citep{llm-as-a-judge}, another leading benchmark for assessing instruction-tuned LLMs' general and multi-turn ability.
It contains a set of challenging multi-turn open-ended questions covering topics such as writing, role-playing, math, coding, and more.
The model-generated answers are judged by GPT-4 via a scalar score (from 1 to 10).

In Table~\ref{tab:extension_model}, we show that \textsc{ExPO} consistently improves the evaluated LLMs, with notable improvements of up to 10.1\% win rate and 4.5\% LC win rate on AlpacaEval 2.0 (for \texttt{internlm2-20b} and \texttt{tulu2-70b}, respectively) and 0.37 on MT-Bench (for \texttt{llama3-8b-iter}).
This suggests that \textit{existing, already-aligned LLMs may still not have been trained to optimality or ``saturation''}.
\textsc{ExPO} offers a practical and efficient means to compensate for potential inadequate training of existing LLMs (or, squeeze more alignment performance out of these models), as it only requires inference-level hardware resources and bypasses the costly additional training overhead.

\begin{table*}[t]
  \centering
  \caption{Evaluation results on UntraFeedback of applying \textsc{ExPO} to models trained via different algorithms.
  }
  \scalebox{0.9}{
    \begin{tabular}{lcccccc}
    \toprule
    & \multicolumn{3}{c}{$\M_0$ is SFTed from Mistral} & \multicolumn{3}{c}{$\M_0$ is SFTed from LLaMA-3} \\
    \cmidrule(lr){2-4}\cmidrule(lr){5-7}
    & \multicolumn{1}{c}{\textbf{Original ($\M_1$)}} & \multicolumn{2}{c}{\textbf{+ \textsc{ExPO} ($\M_2$)}} & \multicolumn{1}{c}{\textbf{Original ($\M_1$)}} & \multicolumn{2}{c}{\textbf{+ \textsc{ExPO} ($\M_2$)}} \\
    \cmidrule(lr){2-2}\cmidrule(lr){3-4}\cmidrule(lr){5-5}\cmidrule(lr){6-7}
    & \textbf{Reward} & \textbf{Optimal $\alpha$} & \textbf{Reward} & \textbf{Reward} & \textbf{Optimal $\alpha$} & \textbf{Reward} \\
    \midrule
    SFT ($\M_0$) & 2.97  & -  & -  & 1.93  & -  & - \\
    \cmidrule{1-7}
    RRHF  & 4.71  & 0.1  & 4.73  \textcolor{red}{\normalsize (+0.02)} & 3.02  & 0.5  & 3.15  \textcolor{red}{\normalsize (+0.13)} \\
    SLiC-HF  & 4.90  & 0.4  & 5.16  \textcolor{red}{\normalsize (+0.26)} & 4.06  & 0.5  & 4.68  \textcolor{red}{\normalsize (+0.62)} \\
    IPO  & 4.97  & 0.5  & 5.44 \textcolor{red}{\normalsize (+0.47)} & 4.75  & 0.3  & 4.86  \textcolor{red}{\normalsize (+0.11)} \\
    CPO  & 4.86  & 0.3  & 5.01 \textcolor{red}{\normalsize (+0.15)} & 4.04  & 0.5  & 4.75  \textcolor{red}{\normalsize (+0.71)} \\
    KTO  & 3.84  & N/A & N/A & 4.48  & 0.4  & 4.67  \textcolor{red}{\normalsize (+0.19)} \\
    R-DPO  & 5.53  & 0.3  & 5.73 \textcolor{red}{\normalsize (+0.20)}  & 4.25  & 0.5  & 4.64  \textcolor{red}{\normalsize (+0.39)} \\
    SimPO  & 5.88  & 0.1  & 5.95 \textcolor{red}{\normalsize (+0.07)}  & 4.89  & 0.4  & 5.21  \textcolor{red}{\normalsize (+0.32)} \\
     \bottomrule
    \end{tabular}%
  }
  \label{tab:extension_algorithm}%
\end{table*}%

\subsection{Applying \textsc{ExPO} to More Alignment Algorithms}
\label{subsec:extension_algorithm}

So far, we have primarily applied \textsc{ExPO} to models trained via the dominant DPO or RLHF algorithms (\S~\ref{sec:experiments} and~\ref{subsec:extension_model}).
Since \textsc{ExPO} does not assume the specific training method for $\M_1$, we expect that \textit{\textsc{ExPO} can be applied to models trained via other algorithms than DPO or RLHF}.
To this end, we use a series of \href{https://huggingface.co/princeton-nlp?search_models=+Mistral-7B-Base-SFT}{Mistral}/\href{https://huggingface.co/princeton-nlp?search_models=+Llama-3-Base-8B-SFT}{LLaMA-3} models released by \citet{simpo}, which are trained via various alignment algorithms and are all initialized from the same SFT checkpoints.
These algorithms include: \textbf{RRHF} \citep{rrhf}, 
\textbf{SLiC-HF} \citep{slic-hf}, \textbf{IPO} \citep{ipo}, \textbf{CPO} \citep{cpo}, \textbf{KTO} \citep{kto}, \textbf{R-DPO} \citep{rdpo}, and \textbf{SimPO} \citep{simpo}.
We refer readers to \citet{simpo} for elaboration on these algorithms' optimization objectives as well as the models' training configurations.
Following the previous experience, we search the optimal $\alpha$ value within the range of [0.1, 0.5] with the interval of 0.1.

As shown in Table~\ref{tab:extension_algorithm}, \textsc{ExPO} effectively complements various alignment training algorithms.
While these models have been carefully tuned according to \citet{simpo}, they still benefit from model extrapolation.
This indicates that \textsc{ExPO} does not rely on specific alignment algorithms but instead generalizes across diverse methods, showcasing its broad compatibility and practical utility.

\subsection{Discussion on Failure Cases}
\label{subsec:failure}

Finally, we discuss the failure cases we encountered when applying \textsc{ExPO} to more various models.
(1) \textsc{ExPO} supposes $\M_0$ is an SFT model and $\M_1$ is one that further undergoes alignment training.
However, when we attempted with a pre-trained model as $\M_0$ and an SFT one as $\M_1$, we found that model extrapolation usually cannot improve alignment performance and can even lead to model collapse (e.g., the extrapolated model struggles to generate the EOS token or mistakenly generates special tokens).
We speculate that this is because SFT typically adopts a larger learning rate and more training steps, and serves to adapt models to the chat templates \citep{chat-templates}, so new knowledge is actually injected into models.
(2) Another type of failure cases is also related to model overfitting.
For example, the \href{https://huggingface.co/jieliu/Storm-7B}{Storm-7B} model \citep{liu2024iterative} is trained via iterative DPO for three iterations.
When experimenting with this model, we found that applying \textsc{ExPO} with even the very small $\alpha=0.1$ results in severe model collapse, probably because the model overfits to its employed reward model during iterative DPO training.

In both cases, \textsc{ExPO}'s underlying first-order approximation can become invalidated as the resulting $\left\|\dtheta\right\|$ is too large.
Therefore, we suggest that more deliberate strategies are needed when applying \textsc{ExPO} to models with large parameter changes, e.g., by leveraging the intermediate checkpoints.
We note that recent work has made promising exploration \citep{lin2025extrapolation} and expect more follow-up studies in future work.

\section{Conclusion}
\label{sec:conclusion}

This work demonstrates the efficacy of the \textsc{ExPO} (model extrapolation) method in enabling more efficient LLM alignment with human preferences.
\textsc{ExPO} builds upon the hypothesis that alignment training typically involves only small changes of model parameters.
Given a partially-trained model $\M_1$ and its initial SFT checkpoint, \textsc{ExPO} improves the implicit optimization objective of alignment training by simply amplifying the parameter change based on a first-order approximation, thus directly achieving better alignment performance without additional training overhead.
We empirically validate \textsc{ExPO}'s effectiveness through controlled experiments, where we show that the DPO model trained with 20\% steps can be boosted to outperform the fully-trained one.
Furthermore, we extend \textsc{ExPO}'s application to twelve existing, already-aligned LLMs, showing that \textsc{ExPO} consistently improves their performance on the mainstream LLM benchmarks AlpacaEval 2.0 and MT-Bench.
This suggests that \textsc{ExPO} can also serve as a practical and efficient means to compensate for potential inadequate alignment training of existing LLMs.
Overall, our work highlights the utility of model extrapolation in efficient LLM alignment, which can inspire future research in this direction.

\section{Limitations}
\label{sec:limitations}

\paragraph{Hyperparameter Search}
The current \textsc{ExPO} adopts the simplest form of uniform extrapolation and requires manual hyperparameter search for $\alpha$.
Future work could explore how to determine the optimal $\alpha$ automatically and adaptively (i.e., using different $\alpha$ values for different model modules).
For example, the information from optimizer states and parameter gradients during the later phase of alignment training could be useful for this purpose.

\paragraph{Alignment Tax}
While \textsc{ExPO} makes substantial improvements in instruction-following ability and alignment with human preferences, this seems not ``free'' and can instead incur an additional \textit{alignment tax}, a widely observed issue in human preference optimization algorithms \citep{instructgpt, rlhflow, simpo}, which indicates the possible fluctuations or drops in downstream task performance after alignment training.
We evaluate the models in \S~\ref{subsec:varying_steps} and~\ref{subsec:extension_model} on the six downstream tasks \citep{arc, hellaswag, mmlu, truthfulqa, winogrande, gsm8k} from the Open LLM Leaderboard\footnote{
We employ the evaluation implementation of Eleuther's \href{https://github.com/EleutherAI/lm-evaluation-harness}{lm-evaluation-harness} (version 0.4.4).
Note that the mismatch of input templates used for chat-style evaluations (e.g., AlpacaEval 2.0 and MT-Bench) and for these downstream task evaluations could also contribute to the observed alignment tax, as discussed in \citet{simpo}.
} (v1; \citealt{open-llm-leaderboard}).
We find that in most cases, \textsc{ExPO} amplifies the alignment tax introduced by the alignment training (from $\M_0$ to $\M_1$).
For example, for the partially-trained models in \S~\ref{subsec:varying_steps} (Appendix~\ref{sec:supp_limitation}, Figure~\ref{fig:lmeval}), the original DPO models ($\M_1$) show improvements over the initial SFT model ($\M_0$) on TruthfulQA and declines on GSM8K, while applying \textsc{ExPO} ($\M_2$) leads to further improvements or declines, respectively.
For the existing, already-aligned LLMs in \S~\ref{subsec:extension_model}, the amplification of the alignment tax by \textsc{ExPO} is usually smaller as shown in Figure~\ref{fig:lmeval_model} in Appendix~\ref{sec:supp_limitation}, suggesting a trade-off between the alignment training overhead (from $\M_0$ to $\M_1$) and the additional alignment tax brought by \textsc{ExPO} (from $\M_1$ to $\M_2$).

\section*{Acknowledgements}
 
We thank Sidi Lu, Yufei Tian, Zi-Yi Dou, and other members of the UCLA PlusLab \& NLP group as well as anonymous reviewers for their constructive feedback and discussions.

This work was supported by an Amazon AGI Research Award through UCLA-Amazon Science Hub and a National
Science Foundation CAREER award \#2339766.
This work was also supported by the National Science Foundation for Distinguished Young Scholars (with No. 62125604) and China Scholarship Council (with No. 202306210211).

\bibliography{custom}

\onecolumn

\clearpage 
\appendix

\section{Related Work}
\label{sec:related}

\paragraph{LLM Alignment}
Modern large language models (LLMs) are first pre-trained on massive textual corpora with the unsupervised language modeling objective \citep{gpt3, llama2, llama3}, and then fine-tuned to learn to follow human instructions \citep{chatgpt, gpt4, ji2023ai}.
The current fine-tuning paradigm typically contains two steps: supervised fine-tuning (SFT) and \textit{human preference optimization}.
Our work focuses on the later step, which aims to adjust the model's response distribution to better \textit{align with human preferences}.
In this process, the model is usually trained on preference data (``A is better than B''; \citealt{slic, click}), thus learning to assign higher probabilities to human-preferred responses over the disfavored ones.
Common implementations for human preference optimization include Reinforcement Learning from Human Feedback (RLHF; \citealt{instructgpt, ppo}), Direct Preference Optimization (DPO; \citealt{dpo}), and many other DPO's variants or competitors \citep{ipo, cpo, kto, rdpo, simpo}.
Given LLMs' gigantic parameters, the processes from pre-training to SFT and the alignment training still require expensive computational resources. 
Therefore, exploring more efficient alignment methods to reduce training overhead has always been an important and compelling research challenge \citep{ji2024aligner}.
To address this challenge, we propose the \textsc{ExPO} method, which has demonstrated promising efficacy in expediting LLM alignment.

There is another line of work that attempts to bypass the expensive alignment training by blending multiple models' token predictions during the inference time \citep{dexperts, contrastive-modeling, realignment}, usually referred to as \textit{inference-time alignment methods}.
In comparison to \textsc{ExPO}, these inference-time methods often require more complex and varied implementations of model inference, which are not typically supported by existing high-performance LLM inference infrastructures (e.g., vLLM).
This inconvenience not only reduces the practical efficiency of model inference but also significantly increases the cost of their hyperparameter search processes.
In contrast, \textsc{ExPO} only involves regular inference of a single model, which can be seamlessly supported by existing infrastructures, thereby inheriting the merit in inference efficiency.

\paragraph{Model Averaging/Interpolation}
Model averaging/interpolation is a commonly used technique in machine learning.
It utilizes multiple models trained with different random initializations or data subsets and interpolates the weights of these models to obtain a new model with stronger out-of-distribution generalization \citep{average-generalization, spurious-feature, model-soups, mitigating-alignment-tax}.
This technique is based on the mode connectivity of neural networks \citep{mode-connectivity, linear-mode-connectivity, zhao2020bridging, frankle2020linear}.
Specifically, prior work found that multiple local optima in the parameter space can often be connected by low-loss (linear) paths, particularly for models with residual connection structures \citep{resnet}.
This can explain why model interpolation can produce new, functional models when applied to LLMs (as our observations in Figure~\ref{fig:interpolation}), as residual connection has become a dominant choice of architecture design in modern LLMs like LLaMA \citep{llama}.
We notice that recent LLMs have widely adopted model interpolation, as exemplified by Gemma-2 \citep{gemma2} and LLaMA-3 \citep{llama3}, possibly also for further enhancement in out-of-distribution generalization.

\clearpage

\section{Hyperparameter Search Details}
\label{sec:hyper_search}

\noindent We use the experiments in Table~\ref{tab:varying_steps} as an example to illustrate how we conduct hyperparameter search.

\paragraph{Starting with $\M_2^{10\%}$:}
(1) First, with an interval of 5, we tried $\alpha=5$ and $\alpha=10$. We found that both significantly outperformed $\M_1$, but $(\alpha=5)>(\alpha=10)$.
(2) Then, setting the search range to $[5, 10]$ with an interval of 1, we applied binary search and tried $\alpha=7$ and $\alpha=8$. We found that $(\alpha=8)>(\alpha=7)$. We then tried $\alpha=9$ and found $(\alpha=8)>(\alpha=9)$.
(3) We thus determined $\alpha=8$ as optimal.

Note that smaller search intervals might yield better results, but we deem this unnecessary in practice.

\paragraph{Then, for $\M_2^{20\%}$:}
(1) With previous experience, we first tried $\alpha=2$ and $\alpha=4$ with an interval of 2. We found that $\alpha=2$ significantly outperformed $\M_1$, but $\alpha=4$ performed worse than $\M_1$.
(2) Then, setting search ranges to $[1, 2]$ and $[2, 4]$ with an interval of 1, we applied binary search and tried $\alpha=1$ and $\alpha=3$. We found that $(\alpha=2)>(\alpha=3)>(\alpha=1)$.
(3) Next, with an interval of 0.5 in $[2, 3]$, we tried $\alpha=2.5$ and found $(\alpha=2.5)>(\alpha=2)$.
(4) We thus determined $\alpha=2.5$ as optimal.

This took 5 searches in total, each taking about 5min (using one A100 80GB, including inference on development set and reward model scoring), totaling about 0.5 GPU hours.

\paragraph{Next, for $\M_2^{40\%}$:}
(1) Based on previous experience, we first tried $\alpha=0.5$ and found it outperformed $\M_0$.
(2) Then with an interval of 0.1, we applied grid search and tried $\alpha=0.6$ and $\alpha=0.4$. We found that $\alpha=0.6$ performed worse than $\M_1$, while $(\alpha=0.5)>(\alpha=0.4)$.
(3) We thus determined $\alpha=0.5$ as optimal.

Note that the search experience for $\M_2^{40\%}$ is a key motivation for us to use [0.1, 0.5] as search range with 0.1 interval for $\M_2^{100\%}$ and models in \S~\ref{subsec:extension_model}.

\paragraph{Summary}
Overall, we (and in practice) do not search blindly, but flexibly combine binary search, grid search, and dynamically adjusted search intervals. 
\textbf{These strategies are simple, practical, and represent consensus in practice.}
It is also noteworthy that the above search only requires \textbf{inference-level GPU hardware} (e.g., A10 24GB).
Therefore, compared to the reduced training overhead (from 12 GPU hours for $\M_1^{100\%}$ to 2.5 GPU hours for $\M_1^{20\%}$) and training-level GPU hardware (from eight A100 80GB to one A10 24GB), the $\alpha$ search process in \textsc{ExPO} is more economical and efficient.

\begin{table}[htbp]
  \centering
  \caption{
  Hyperparameter search results for $\alpha$ in \S~\ref{subsec:varying_steps} and~\ref{subsec:extension_model}.
  }
  \scalebox{0.9}{
    \begin{tabular}{llcc}
    \toprule
    &  & \textbf{Search Interval} & \textbf{Optimal $\alpha$} \\
   \midrule
   \multirow{4}[0]{*}{\tabincell{l}{\textbf{Models in \S~\ref{subsec:varying_steps}} \\ (binary/grid search)}} & DPO (10\% data) & 1.0 & 8.0 \\
   & DPO (20\% data)  & 0.5 & 2.5 \\
   & DPO (40\% data)  & 0.1 & 0.5 \\
   & \texttt{zephyr-7b-dpo}  & 0.1 & 0.3 \\
    \midrule
    \multirow{6}[0]{*}{\tabincell{l}{\textbf{Models in \S~\ref{subsec:extension_model}} \\ (grid search within [0.1, 0.5])}} & \texttt{zephyr-7b-alpha/beta} & 0.1 & 0.3/0.1 \\
    & \texttt{tulu2-7/13/70b} & 0.1 & 0.5 \\
    & \texttt{snorkel-7b-iter} & 0.1 & 0.3 \\
    & \texttt{llama3-8b-iter} & 0.1 & 0.3 \\
    & \texttt{starling-7b-alpha/beta} & 0.1 & 0.2/0.5 \\
    & \texttt{internlm2-1.8/7/20b} & 0.1 & 0.5 \\
    \bottomrule
    \end{tabular}%
  }
  \label{tab:alpha}%
\end{table}%

\clearpage

\section{HuggingFace Models}
\label{sec:models}

\begin{table}[htbp]
    \centering
    \scalebox{0.9}{
    \begin{tabular}{p{5cm}cl}
    \toprule
     &  & \textbf{HuggingFace Model ID} \\
    \midrule
       \multirow{2}[0]{*}{Reward models} & & \texttt{weqweasdas/RM-Mistral-7B} \\
       & & \texttt{sfairXC/FsfairX-LLaMA3-RM-v0.1} \\
   \midrule
     \multirow{2}[0]{*}{\texttt{zephyr-7b-dpo}} & $\M_0$ & \texttt{alignment-handbook/zephyr-7b-sft-full} \\
      & $\M_1$ & \texttt{alignment-handbook/zephyr-7b-dpo-full} \\
      \midrule
    \multirow{2}[0]{*}{\texttt{zephyr-7b-\{alpha/beta\}}} & $\M_0$ & \texttt{HuggingFaceH4/mistral-7b-sft-\{alpha/beta\}} \\
      & $\M_1$ & \texttt{HuggingFaceH4/zephyr-7b-\{alpha/beta\}} \\
      \midrule[0mm]
      \multirow{2}[0]{*}{\texttt{tulu2-\{7/13/70\}b}} & $\M_0$ & \texttt{allenai/tulu-2-\{7/13/70\}b} \\
      & $\M_1$ & \texttt{allenai/tulu-2-dpo-\{7/13/70\}b} \\
      \midrule[0mm]
      \multirow{2}[0]{*}{\texttt{snorkel-7b-iter}} & $\M_0$ & \texttt{mistralai/Mistral-7B-Instruct-v0.2} \\
      & $\M_1$ & \texttt{snorkelai/Snorkel-Mistral-PairRM-DPO} \\
      \midrule[0mm]
      \multirow{2}[0]{*}{\texttt{llama3-8b-iter}} & $\M_0$ & \texttt{RLHFlow/LLaMA3-SFT} \\
      & $\M_1$ & \texttt{RLHFlow/LLaMA3-iterative-DPO-final} \\
      \midrule[0mm]
      \multirow{2}[0]{*}{\texttt{starling-7b-alpha}} & $\M_0$ & \texttt{openchat/openchat\_3.5} \\
      & $\M_1$ & \texttt{berkeley-nest/Starling-LM-7B-alpha} \\
      \midrule[0mm]
      \multirow{2}[0]{*}{\texttt{starling-7b-beta}} & $\M_0$ & \texttt{openchat/openchat-3.5-0106} \\
      & $\M_1$ & \texttt{Nexusflow/Starling-LM-7B-beta} \\
      \midrule[0mm]
      \multirow{2}[0]{*}{\texttt{internlm2-\{1.8/7/20\}b}} & $\M_0$ & \texttt{internlm/internlm2-chat-\{1\_8/7/20\}b-sft} \\
      & $\M_1$ & \texttt{internlm/internlm2-chat-\{1\_8/7/20\}b} \\
      \midrule
      Mistral-based SFT & $\M_0$ & \texttt{alignment-handbook/zephyr-7b-sft-full} \\
      \{RRHF, SLiC-HF, IPO, CPO, KTO, R-DPO, SimPO\} & \multirow{2}[0]{*}{$\M_1$} & \multirow{2}[0]{*}{\texttt{princeton-nlp/Mistral-7B-Base-SFT-\{*\}}} \\
      \midrule
      LLaMA-3-based SFT & $\M_0$ & \texttt{princeton-nlp/Llama-3-Base-8B-SFT} \\
      \{RRHF, SLiC-HF, IPO, CPO, KTO, R-DPO, SimPO\} & \multirow{2}[0]{*}{$\M_1$} & \multirow{2}[0]{*}{\texttt{princeton-nlp/Llama-3-Base-8B-SFT-\{*\}}} \\
    \bottomrule
    \end{tabular}
    }
\end{table}

\section{Supplementary Experimental Results of Alignment Tax (\S~\ref{sec:limitations})}
\label{sec:supp_limitation}

\begin{figure}[!h]
  \centering
  \includegraphics[width=\linewidth]{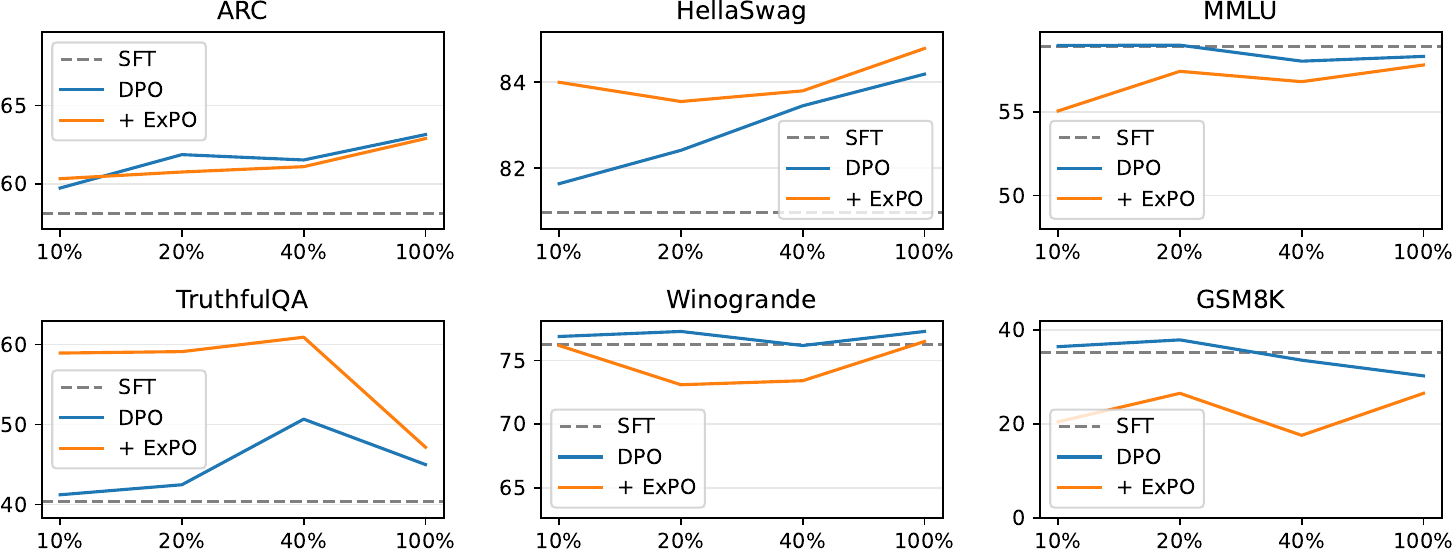}
  \caption{
  Evaluation results for the models in \S~\ref{subsec:varying_steps} on downstream tasks.
  The x-axis denotes the proportions of training steps.
  As the ``cost'' of simply improving instruction-following ability and alignment with human preferences, \textsc{ExPO} can also amplify the alignment tax introduced by the alignment training.
  }
  \label{fig:lmeval}
\end{figure}

\begin{figure}[htbp]
  \centering
  \includegraphics[width=\linewidth]{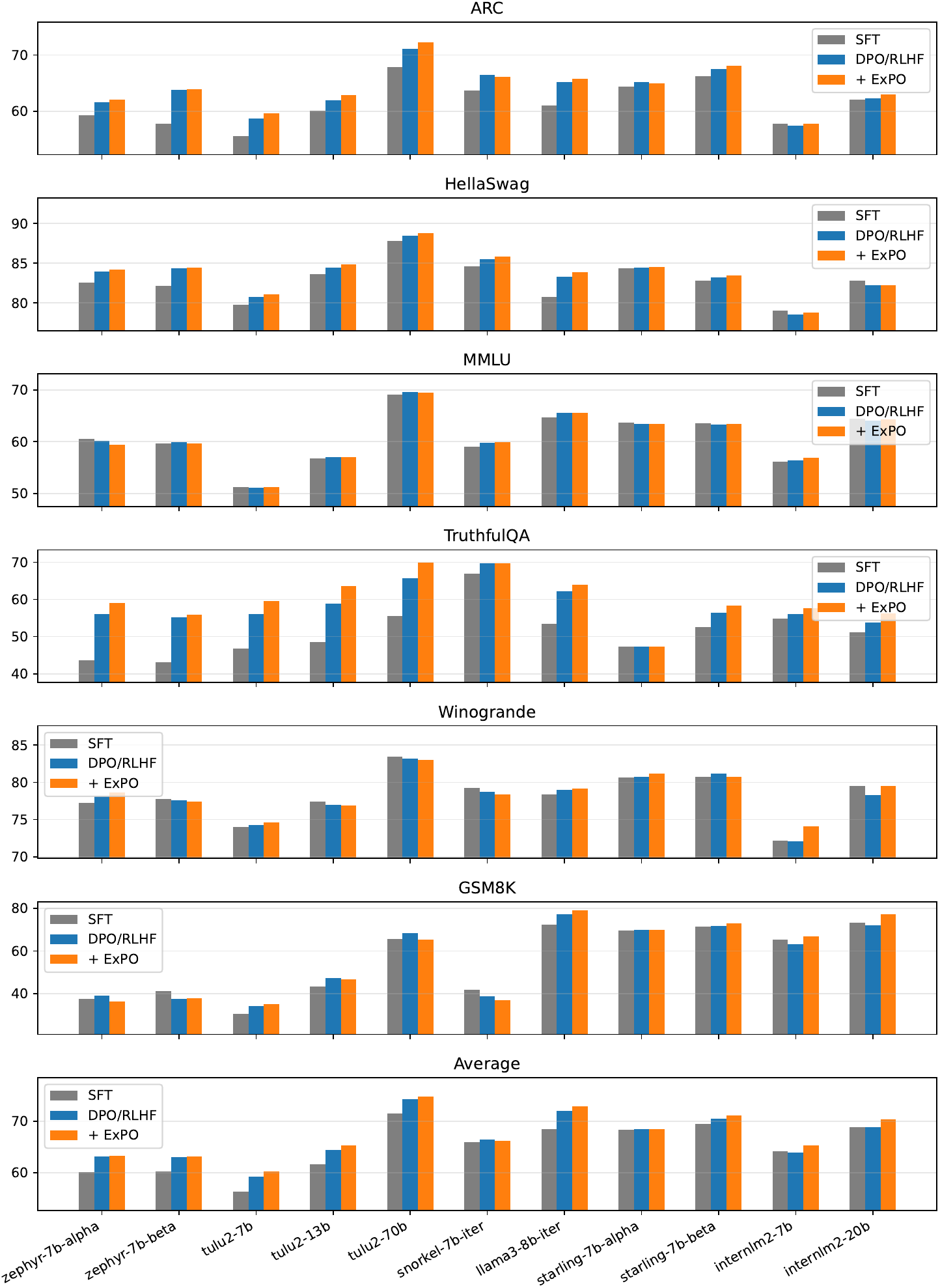}
  \caption{
  Evaluation results for the LLMs in \S~\ref{subsec:extension_model} on downstream tasks.
  For these already-alighed models, the additional alignment tax brought by \textsc{ExPO} is usually smaller, suggesting a trade-off between the alignment training overhead (from $\M_0$ to $\M_1$) and the additional alignment tax brought by \textsc{ExPO} (from $\M_1$ to $\M_2$).
  }
  \label{fig:lmeval_model}
\end{figure}

\end{document}